\pdfoutput=1

\documentclass[11pt]{article}

\usepackage{TrustNLP2023}

\usepackage{times}
\usepackage{latexsym}

\usepackage[T1]{fontenc}

\usepackage[utf8]{inputenc}

\usepackage{microtype}

\usepackage{inconsolata}

\usepackage[ruled,linesnumbered]{algorithm2e}
\usepackage[justification=centering]{caption}
\usepackage{subcaption}
\usepackage{booktabs}
\usepackage{amsmath}
\usepackage{makecell}
\usepackage{graphicx}
\usepackage{tabularx}
\usepackage{stfloats}
\usepackage{pfnote}

\title{Pay Attention to the Robustness of Chinese Minority Language Models! Syllable-level Textual Adversarial Attack on Tibetan Script}

\makeatletter
\def\thanks#1{\protected@xdef\@thanks{\@thanks\protect\footnotetext{#1}}}
\makeatother

\author{Xi Cao\textsuperscript{1, 2}, Dolma Dawa\textsuperscript{1, 2}, Nuo Qun\textsuperscript{1, 2 *}\thanks{* Corresponding author.}, Trashi Nyima\textsuperscript{1, 2} \\
	\textsuperscript{1}School of Information Science and Technology, \\Tibet University, Lhasa, Tibet 850000, China \\
	\textsuperscript{2}Collaborative Innovation Center for Tibet Informatization by \\MOE and Tibet Autonomous Region, Lhasa, Tibet 850000, China \\
	{\tt metaphor@outlook.com}, {\tt \{da\_zhui,q\_nuo,nmzx\}@utibet.edu.cn} \\}

\begin{document}
\maketitle
\begin{abstract}
The textual adversarial attack refers to an attack method in which the attacker adds imperceptible perturbations to the original texts by elaborate design so that the NLP (natural language processing) model produces false judgments.
This method is also used to evaluate the robustness of NLP models.
Currently, most of the research in this field focuses on English, and there is also a certain amount of research on Chinese.
However, to the best of our knowledge, there is little research targeting Chinese minority languages.
Textual adversarial attacks are a new challenge for the information processing of Chinese minority languages.
In response to this situation, we propose a Tibetan syllable-level black-box textual adversarial attack called TSAttacker based on syllable cosine distance and scoring mechanism.
And then, we conduct TSAttacker on six models generated by fine-tuning two PLMs (pre-trained language models) for three downstream tasks.
The experiment results show that TSAttacker is effective and generates high-quality adversarial samples.
In addition, the robustness of the involved models still has much room for improvement.
\end{abstract}

\section{Introduction}


With the development of neural network models, methods based on the models have been widely used in many fields and achieved remarkable performance, such as computer vision, speech recognition, and natural language processing.
However, neural network models are susceptible to adversarial attacks \cite{szegedy-etal-2013-intriguing}.

When textual adversarial attacks are performed on the NLP models for classification tasks, the models with high robustness will make predictions consistent with the original texts after perturbation, while the models with low robustness will make incorrect predictions.
Section \ref{Section2} will detail the current research status of textual adversarial attacks on English and Chinese.
The information processing technology of Chinese minority languages started late, but in recent years, the emergence of Chinese minority PLMs has promoted development but brought new challenges, one of which is textual adversarial attacks.
However, there is little research on this topic.

The main contributions of this paper are as follows:

(1) To fill the research gap of textual adversarial attacks on Tibetan script, this paper proposes TSAttacker, a Tibetan syllable-level black-box textual adversarial attack with a high attack success rate.
This attack method can significantly reduce the accuracy of the models and generate adversarial samples with a low average Levenshtein distance.

(2) To evaluate the robustness of the Tibetan part in the first Chinese minority multilingual PLM, this paper conducts TSAttacker on six models generated by fine-tuning two versions of the PLM for three downstream tasks.
During fine-tuning, we also find that training sets conforming to language standards can improve model performance.

(3) To facilitate future explorations, we open-source our work on GitHub (\url{https://github.com/UTibetNLP/TSAttacker}).
We call on more researchers to pay attention to the security issues in the information processing of Chinese minority languages.

\section{Related Work}
\label{Section2}

\subsection{Textual Adversarial Attacks on English}

At present, most of the research on textual adversarial attacks concentrates on English.
\citet{jia-liang-2017-adversarial} first proposed generating adversarial samples for English public datasets and evaluating NLP models from a robustness perspective.
Since then, various English textual adversarial attack methods with different strategies have emerged.
According to the granularity of text perturbations, attacks can be classified into character-, word-, and sentence-level \cite{du-etal-2021-adversarial}.

Character-level attacks are operations that perturb the characters of the original text, including adding, deleting, modifying, and changing the order of characters.
\citet{ebrahimi-etal-2018-hotflip} proposed a character-level white-box attack method called HotFlip based on the gradients of the one-hot input vectors, \citet{gao-etal-2018-black} proposed a greedy algorithm based on scoring called DeepWordBug for character-level black-box attacks, \citet{eger-etal-2019-text} proposed a character-level white-box attack method called VIPER based on visual similarity, and so on.

Word-level attacks are to perturb the words of the original text, and the main method is word substitution.
Such as, \citet{jin-etal-2020-is} proposed a word-level black-box attack method called TextFooler which combines the cosine similarity of words with the semantic similarity of sentences, \citet{garg-ramakrishnan-2020-bae} proposed a word-level black-box attack method based on the BERT mask language model called BAE, and \citet{choi-etal-2022-tabs} proposed TABS, an efficient beam search word-level black-box attack method.

Sentence-level attacks generate adversarial sentences primarily through paraphrasing and text generation, which often result in a significant gap between the perturbed text and the original text.
Moreover, it is difficult to control the quality of generated adversarial samples.
The attack effect is also relatively average \cite{zheng-etal-2021-survey}.

\subsection{Textual Adversarial Attacks on Chinese}

The methods of generating adversarial texts are closely related to language characteristics, such as textual features and grammatical structure.
Therefore, there are different methods of generating adversarial samples for various languages.
The research on Chinese textual adversarial attacks started later than English, but there are also some related studies.
\citet{wang-etal-2019-adversarial} proposed a Chinese word-level black-box attack method called WordHanding, which designed a new word importance calculation algorithm and utilized homonym substitution to generate adversarial samples.
\citet{tong-etal-2020-a} proposed a Chinese word-level black-box attack method called CWordAttacker, which used the targeted deletion scoring mechanism and substituted words with traditional Chinese and Pinyin.
\citet{zhang-etal-2022-character} proposed a Chinese character-level black-box attack method called PGAS, which generated adversarial samples with minor disturbance by replacing polyphonic characters.
The relevant research on Chinese textual adversarial attacks is not sufficient, and the language features of Chinese are not fully integrated.
So, there is still a lot of exploration space.

\subsection{Textual Adversarial Attacks on Chinese Minority Languages}

With the construction and development of information technology in Chinese minority areas like Inner Mongolia, Tibet, and Xinjiang, the corpus of Chinese minority languages has reached a certain scale.
Recently, there have been some PLMs targeting or containing Chinese minority languages.
It is worth mentioning that \citet{yang-etal-2022-cino} proposed CINO (Chinese mINOrity PLM), the first Chinese minority multilingual PLM, covering standard Chinese, Cantonese, Tibetan, Mongolian, Uyghur, Kazakh, Zhuang, and Korean.
This model achieves SOTA (state-of-the-art) performance on multiple monolingual or multilingual datasets for text classification, significantly promoting the NLP research of Chinese minority languages.

Meanwhile, \citet{morris-etal-2020-textattack} released an English textual adversarial attack frame called TextAttack, \citet{zeng-etal-2021-openattack} released a textual adversarial attack toolkit called OpenAttack which supports both English and Chinese, 
\citet{wang-etal-2021-textflint} released a robustness evaluation toolkit called TextFlint for English and Chinese NLP models, etc.
These have provided a good research platform for other languages' textual adversarial attacks and model robustness evaluation.

However, to the best of our knowledge, there is little research involving textual adversarial attacks on Chinese minority languages such as Mongolian, Tibetan, and Uyghur.
Without robustness evaluation, the NLP models with low robustness will face serious risks, such as hacker attacks, poor user experience, and political security problems, which pose a huge threat to the stable development and information construction of Chinese minority areas.
Therefore, we should take precautions to study the textual adversarial attack methods of related languages and evaluate the robustness of related models to fill in the gaps in related research fields.

\section{Attack Method}

\subsection{Textual Adversarial Attacks on Text Classification}

For a $K$-class classification dataset $D=\{(x_{i}, y_{i})\}_{i=1}^{n}$, where $x\in{X}$ ($X$ includes all possible input texts) and $y\in{Y}$ ($Y$ includes all $K$ classifications). 
The classifier $F$ obtains the classification $y_{true}$ corresponding to the original input text $x$, denoted as
\begin{equation}
	\label{Equation1}
	F(x)=arg\max_{y\in{Y}}P(y|x)={y_{true}}.
\end{equation}
The attacker achieves a successful textual adversarial attack by elaborately designing the adversarial text $x'$ and making
\begin{equation}
	\label{Equation2}
	F(x')=arg\max_{y\in{Y}}P(y|x')\neq{y_{true}},
\end{equation}
where $x'$ is the result of adding $\epsilon$-bounded, imperceptible perturbations $\delta$ to the original text $x$.

\subsection{TSAttacker Algorithm}

Tibetan is a phonetic script consisting of 30 consonant letters and 4 vowel letters. 
These letters are combined into Tibetan syllables according to certain rules. 
A Tibetan word is composed of one or more syllables separated by tsheg (\includegraphics[scale=0.2]{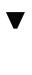}).
Therefore, it is different from English and Chinese in that the syllable granularity in Tibetan is between character and word.
Let the syllable in the original input text $x$ be $s$ (ignore tsheg and end-of-sentence punctuation), then
\begin{equation}
	\label{Equation3}
	x=s_{1}s_{2}\dots{s_{i}}\dots{s_{n}}.
\end{equation}

In this work, we propose a Tibetan syllable-level black-box textual adversarial attack called TSAttacker based on syllable cosine distance and scoring mechanism.
We adopt syllable cosine distance to obtain syllables for substitution and a scoring mechanism to determine the order of syllable substitutions.
The pseudocode of the TSAttacker algorithm is as shown in Appendix \ref{AppendixA}.

\subsubsection{Syllable Substitution}

\citet{grave-etal-2018-learning} released high-quality pre-trained word vectors for 157 languages, including Tibetan syllable embeddings, which were trained using fastText\footnote{\url{https://fasttext.cc}} \cite{joulin-etal-2016-fasttext} on the dataset composed of a mixture of Wikipedia and Common Crawl.
The Tibetan training result contains some unwanted vectors due to the nature of the training dataset, such as embeddings of ``MP3'', ``PNG'', and ``File''.
Consequently, we cleaned the result and obtained 7,652 Tibetan syllable embeddings basically containing all commonly used syllables.

For each Tibetan syllable $s$ in the original input text $x$, we use all syllables whose embedding's cosine distances from the embedding of $s$ are within the range of $(0,d_{max}]$ as a candidate syllables' set $C$.
Let the cosine distance between the embedding of $s$ and the embedding of $s'$ ($s'\in{C}$) be $d$, then $d$ satisfies the following condition:
\begin{equation}
	\label{Equation4}
	d={1-\frac{\mathbf{s}\cdot\mathbf{s'}}{|\mathbf{s}|\cdot|\mathbf{s'}|}}\le{d_{max}}.
\end{equation}
By adjusting $d_{max}$, we can control the similarity between all syllables in set $C$ and syllable $s$.
The smaller $d_{max}$ is, the more similar all syllables in set $C$ are to syllable $s$.
As a result, the size of set $C$ can be adjusted.
The larger $d_{max}$, the larger the size of set $C$.

For the $i$-th Tibetan syllable $s_{i}$ in the original input text $x$, there is always a candidate syllables' set $C_{i}$ corresponding to it.
Assuming that the size of set $C_{i}$ is $m$.
We select a candidate syllable ${s_{i}}'$ from set $C_{i}$ each time, and
\begin{equation}
	\label{Equation5}
	{x_{i}}'=s_{1}s_{2}\dots{{s_{i}}'}\dots{s_{n}}.
\end{equation}
At the same time, we calculate
\begin{equation}
	\label{Equation6}
	\Delta{P_{i}}=P(y_{true}|x)-P(y_{true}|{x_{i}}').
\end{equation}
After iterating set $C_{i}$, the syllable ${s_{i}}^{*}$ can be found, and
\begin{equation}
	\label{Equation7}
	{x_{i}}^{*}=s_{1}s_{2}\dots{{s_{i}}^{*}}\dots{s_{n}}.
\end{equation}
At the moment,
\begin{align}
	\label{Equation8}
	\Delta{{P_{i}}^{*}}&=P(y_{true}|x)-P(y_{true}|{x_{i}}^{*}) \\
	&=max\{\Delta{P_{ij}\}}_{j=1}^{m} \notag \\
	&=max\{P(y_{true}|x)-P(y_{true}|{{x_{i}}'}_{j})\}_{j=1}^{m}, \notag
\end{align}
\begin{align}
	\label{Equation9}
	{s_{i}}^{*}&=arg\max_{{{s_{i}}'}\in{C_{i}}}\{\Delta{P_{ij}\}}_{j=1}^{m} \\
	&=arg\max_{{{s_{i}}'}\in{C_{i}}}\{P(y_{true}|x)-P(y_{true}|{{x_{i}}'}_{j})\}_{j=1}^{m}. \notag
\end{align}
The syllable ${s_{i}}^{*}$ obtained in this way can cause the most significant change in the classification probability after substitution and have the best attack effect.
Therefore, we use syllable ${s_{i}}^{*}$ to substitute syllable $s_{i}$.

\subsubsection{Substitution Order}

Word saliency \cite{li-etal-2016-understanding} refers to the degree of change in the classification probability after a word is set to unknown (out of vocabulary).
Here, we use it to calculate syllable saliency.
For the $i$-th Tibetan syllable $s_{i}$ in the original input text $x$, we set it to ``$<UNK>$'', and
\begin{equation}
	\label{Equation10}
	\hat{{x}_{i}}=s_{1}s_{2}\dots{<UNK>}\dots{s_{n}}.
\end{equation}
Then, we calculate the saliency of syllable $s_{i}$ as $S_{i}$:
\begin{equation}
	\label{Equation11}
	S_{i}=P(y_{true}|x)-P(y_{true}|\hat{x_{i}}).
\end{equation}
We incorporate the scoring formula in the probability weighted word saliency algorithm \cite{ren-etal-2019-generating} to determine the substitution order of syllables in the original input text $x$.
The score $H_{i}$ is defined as follows:
\begin{align}
	\label{Equation12}
	H_{i}&=Softmax(S_{i})\cdot\Delta{{P_{i}}^{*}} \\
	&=\frac{e^{S_{i}}}{\Sigma_{j=1}^{n}e^{S_{j}}}\cdot\Delta{{P_{i}}^{*}}. \notag
\end{align}
From the above formula, it can be seen that the score $H_{i}$ comprehensively considers the importance of the substituted syllable $s_{i}$ and the substitution syllable ${s_{i}}^{*}$.
After sorting $n$ scores $\{H_{1},H_{2},\dots,H_{n}\}$ corresponding to the original input text $x$ in descending order, we sequentially substitute $s_{i}$ with ${s_{i}}^{*}$.
If ${F(x')}\neq{F(x)}$, the attack succeeds, and if always $F(x')=F(x)$, the attack fails.

\section{Experiment}

\subsection{Datasets and Models}

\subsubsection{Datasets}

Table \ref{Table1} lists the detailed information of the datasets: TNCC-title, TNCC-document, and TU\_SA, including task, number of classes, the average number of syllables, etc.

\textbf{TNCC}\footnote{\url{https://github.com/FudanNLP/Tibetan-Classification}}.
\citet{qun-etal-2017-end} open-sourced the Tibetan News Classification Corpus (TNCC) collected from the China Tibet Online website (\url{http://tb.tibet.cn}).
This corpus consists of two parts: TNCC-title, a news title classification dataset, and TNCC-document, a news document classification dataset.
TNCC-title is a short text dataset with 9,276 samples and an average of 16 syllables per title.
TNCC-document is a long text dataset with 9,204 samples and an average of 689 syllables per document.
There are twelve classes both in TNCC-title and TNCC-document dataset: Politics, Economics, Education, Tourism, Environment, Language, Literature, Religion, Arts, Medicine, Customs, and Instruments.

\textbf{TU\_SA}\footnote{\url{https://github.com/UTibetNLP/TU_SA}}.
TU\_SA is a Tibetan sentiment classification dataset consisting of 10,000 samples labeled as positive or negative, with 5,000 samples in each class.
\citet{zhu-etal-2023-sentiment} selected 10,000 sentences from the public Chinese sentiment analysis datasets: weibo\_senti\_100k and ChnSentiCorp, then manually translated and proofread by professional researchers to form this dataset.

\subsubsection{Models}

The existing public PLMs targeting or containing Tibetan include the monolingual PLM TiBERT \cite{liu-etal-2022-tibert} based on BERT \cite{devlin-etal-2019-bert} and the multilingual PLM CINO \cite{yang-etal-2022-cino} based on XLM-R \cite{conneau-etal-2020-unsupervised}, and CINO has achieved SOTA performance in relevant evaluations on Tibetan.
We adopt two versions of CINO: cino-base-v2\footnote{\url{https://huggingface.co/hfl/cino-base-v2}} and cino-large-v2\footnote{\url{https://huggingface.co/hfl/cino-large-v2}}, then fine-tune them for the three downstream tasks corresponding to the above datasets.
Each dataset is split into a training set, a validation set, and a test set according to a ratio of 8:1:1.
We select the best checkpoints based on the macro-F1 score for TNCC and the F1 score for TU\_SA.
The hyperparameters used for downstream fine-tuning are listed in Table \ref{Table2}.

It should be noted that the texts in TNCC have been pre-tokenized, which means that a space instead of a tsheg has been added between two syllables.
When \citet{yang-etal-2022-cino} fine-tuned CINO on TNCC, they removed the spaces, but the processed texts do not conform to the standards of Tibetan script, and there should be a tsheg between two syllables.
Therefore, we make a separate experiment that fine-tunes models on texts with a space between two syllables, texts with no space between two syllables, and texts with a tsheg between two syllables.
The results of the validation sets are listed in the first 12 rows of Table \ref{Table3} and show that models fine-tuned on the texts conforming to language standards can achieve better performance.

Table \ref{Table3} list the performance of the models fine-tuned on TNCC and TU\_SA.
We adopt the following six models as victim models and conduct TSAttacker on the test sets: cino-base-v2+TNCC-title(tsheg), cino-base-v2+TNCC-document(tsheg), cino-large-v2+TNCC-title(tsheg), cino-large-v2+TNCC-document(tsheg), cino-base-v2+TU\_SA, and cino-large-v2+TU\_SA.

\begin{table*}
	\centering
	\caption{Detailed information of the datasets.}
	\label{Table1}
	\begin{tabular}{cccccccc}
		\toprule
		Dataset & Task & \#Classes & \makecell{\#Average\\syllables} & \makecell{\#Total\\samples} & \makecell{\#Train\\samples} & \makecell{\#Validation\\samples} & \makecell{\#Test\\samples} \\
		\midrule
		\makecell{TNCC-\\title} & \makecell{news title\\classification} & 12 & 16 & 9,276 & 7,422 & 927 & 927 \\
		\hline
		\makecell{TNCC-\\document} & \makecell{news document\\classification} & 12 & 689 & 9,204 & 7,364 & 920 & 920 \\
		\hline
		TU\_SA & \makecell{sentiment\\classification} & 2 & 28 & 10,000 & 8,000 & 1,000 & 1,000 \\
		\bottomrule
	\end{tabular}
\end{table*}

\begin{table*}
	\centering
	\caption{Hyperparameters used for downstream fine-tuning.}
	\label{Table2}
	\begin{tabular}{cccccc}
		\toprule
		Model & Dataset & Batch size & Epochs & Learning rate & Warmup ratio \\
		\midrule
		cino-base-v2 & TNCC \& TU\_SA & 32 & 40 & 5e-5 & 0.1 \\
		cino-large-v2 & TNCC \& TU\_SA & 32 & 40 & 3e-5 & 0.1 \\
		\bottomrule
	\end{tabular}
\end{table*}

\begin{table*}
	\centering
	\caption{Model performance on TNCC and TU\_SA..}
	\label{Table3}
	\begin{tabularx}{\linewidth}{cccccccc}
		\toprule
		\makecell{Model\\(PLM+Dataset)} & Accuracy & \makecell{Macro-\\F1} & \makecell{Macro-\\Precision} & \makecell{Macro-\\Recall} & \makecell{Weighted\\-F1} & \makecell{Weighted\\-Precision} & \makecell{Weighted\\-Recall} \\
		\midrule
		\makecell{cino-base-v2+\\TNCC-title\\(space)} & 0.6624 & 0.6375 & \textbf{0.6721} & 0.6213 & 0.6564 & 0.6613 & 0.6624 \\
		\hline
		\makecell{cino-base-v2+\\TNCC-title\\(no space)} & 0.6602 & 0.6385 & 0.6382 & 0.6454 & 0.6621 & 0.6716 & 0.6602 \\
		\hline
		\textbf{\makecell{cino-base-v2+\\TNCC-title\\(tsheg)}} & \textbf{0.6764} & \textbf{0.6488} & 0.6523 & \textbf{0.6556} & \textbf{0.6772} & \textbf{0.6853} & \textbf{0.6764} \\
		\hline
		\hline
		\makecell{cino-base-v2+\\TNCC-document\\(space)} & 0.7380 & 0.6985 & 0.7039 & 0.6949 & 0.7382 & 0.7399 & 0.7380 \\
		\hline
		\makecell{cino-base-v2+\\TNCC-document\\(no space)} & 0.7435 & 0.6967 & 0.7241 & 0.6817 & 0.7430 & 0.7501 & 0.7435 \\
		\hline
		\textbf{\makecell{cino-base-v2+\\TNCC-document\\(tsheg)}} & \textbf{0.7598} & \textbf{0.7317} & \textbf{0.7502} & \textbf{0.7180} & \textbf{0.7602} & \textbf{0.7630} & \textbf{0.7598} \\
		\hline
		\hline
		\makecell{cino-large-v2+\\TNCC-title\\(space)} & 0.6785 & 0.6448 & 0.6489 & 0.6449 & 0.6767 & 0.6786 & 0.6785 \\
		\hline
		\makecell{cino-large-v2+\\TNCC-title\\(no space)} & 0.6861 & 0.6568 & 0.6818 & 0.6429 & 0.6831 & 0.6874 & 0.6861 \\
		\hline
		\textbf{\makecell{cino-large-v2+\\TNCC-title\\(tsheg)}} & \textbf{0.7044} & \textbf{0.6759} & \textbf{0.6898} & \textbf{0.6672} & \textbf{0.7025} & \textbf{0.7062} & \textbf{0.7044} \\
		\hline
		\hline
		\makecell{cino-large-v2+\\TNCC-document\\(space)} & 0.7380 & 0.6985 & 0.7039 & 0.6949 & 0.7382 & 0.7399 & 0.7380 \\
		\hline
		\makecell{cino-large-v2+\\TNCC-document\\(no space)} & 0.7435 & 0.6967 & 0.7241 & 0.6817 & 0.7430 & 0.7501 & 0.7435 \\
		\hline
		\textbf{\makecell{cino-large-v2+\\TNCC-document\\(tsheg)}} & \textbf{0.7815} & \textbf{0.7642} & \textbf{0.7749} & \textbf{0.7576} & \textbf{0.7827} & \textbf{0.7862} & \textbf{0.7815} \\
		\hline
		\hline
		\textbf{\makecell{cino-base-v2+\\TU\_SA}} & 0.7530 & \makecell{0.7748\\(F1)} & \makecell{0.7119\\(Precision)} & \makecell{0.8500\\(Recall)} & - & - & - \\
		\hline
		\textbf{\makecell{cino-large-v2+\\TU\_SA}} & 0.7970 & \makecell{0.7992\\(F1)} & \makecell{0.7906\\(Precision)} & \makecell{0.8080\\(Recall)} & - & - & - \\
		\bottomrule
	\end{tabularx}
\end{table*}

\subsection{Evaluation Metrics and Experiment Results}

We use \textit{Accuracy Drop Value} (ADV) and \textit{Attack Success Rate} (ASR) to evaluate both the attack effectiveness and the model robustness, and \textit{Levenshtein Distance} (LD) to evaluate the quality of a generated adversarial sample.
ADV refers to the difference in the model accuracy on the test set between pre-attack and post-attack.
ASR refers to the percentage of the attack that successfully fool the victim model.
The larger ADV or ASR, the more effective the attack and the less robust the model.
LD refers to the minimum number of single-syllable edits between two texts, like insertions, deletions, and substitutions.
The smaller LD, the higher the quality of the generated adversarial sample.

In this work, we set the maximum cosine distance $d_{max}$ to 0.2929, in other words, the maximum angle between two syllable embeddings is $45^{\circ}$.
We use this parameter to determine the set of candidate substitution syllables according to Equation \ref{Equation4}.
Table \ref{Table4} shows the experiment results and Appendix \ref{AppendixB} lists some adversarial samples generated by TSAttacker.

\begin{table*}
	\centering
	\caption{Experiment results.\\ADV = Accuracy Drop Value, ASR = Attack Success Rate, LD = Levenshtein Distance.}
	\label{Table4}
	\begin{tabular}{cccm{1.8cm}<{\centering}m{1.8cm}<{\centering}m{1.8cm}<{\centering}}
		\toprule
		\makecell{Model\\(PLM+Dataset)} & \makecell{Accuracy\\(pre-attack)} & \makecell{Accuracy\\(post-attack)} & ADV ($\uparrow$) & ASR ($\uparrow$) & Average LD ($\downarrow$) \\
		\midrule
		\makecell{cino-base-v2+\\TNCC-title(tsheg)} & 0.6731 & 0.3085 & \underline{0.3646} & \underline{0.7605} & \underline{1.6411} \\
		\hline
		\makecell{cino-large-v2+\\TNCC-title(tsheg)} & 0.6850 & 0.3420 & 0.3430 & 0.7487 & 1.7176 \\
		\hline
		\hline
		\makecell{cino-base-v2+\\TNCC-document(tsheg)} & 0.7576 & 0.3717 & \underline{0.3859} & \underline{0.7120} & \underline{39.1800} \\
		\hline
		\makecell{cino-large-v2+\\TNCC-document(tsheg)} & 0.7500 & 0.4217 & 0.3283 & 0.6696 & 41.9660 \\
		\hline
		\hline
		\makecell{cino-base-v2+\\TU\_SA} & 0.7430 & 0.5190 & 0.2240 & 0.6380 & 2.9404 \\
		\hline
		\makecell{cino-large-v2+\\TU\_SA} & 0.7760 & 0.5100 & \underline{0.2660} & \underline{0.6570} & \underline{2.7017} \\
		\bottomrule
	\end{tabular}
\end{table*}

The results show that our proposed attack method TSAttacker greatly reduces the model accuracy and has a high attack success rate, which shows the effectiveness of the attack method.
For the dataset TNCC-title, the accuracy of the models cino-base-v2 and cino-large-v2 decreases by 0.3646 and 0.3430, and the attack success rate reaches 0.7605 and 0.7487, respectively;
for the dataset TNCC-document, the accuracy of the models cino-base-v2 and cino-large-v2 decreases by 0.3859 and 0.3283, and the attack success rate reaches 0.7120 and 0.6696, respectively;
for the dataset TU\_SA, the accuracy of the models cino-base-v2 and cino-large-v2 decreases by 0.2240 and 0.2660, and the attack success rate reaches 0.6380 and 0.6570, respectively.

From a certain point of view, the robustness of Chinese minority NLP models still has much room for improvement.
The model cino-base-v2 is a base version of CINO, with 12 layers, 768 hidden states, and 12 attention heads.
The model cino-large-v2 is a large version of CINO, with 24 layers, 1024 hidden states, and 16 attention heads.
However, for different datasets, the same attack method does not always achieve a higher attack success rate on the smaller model, and the larger model is not always the one with a smaller accuracy drop value.
This seems to indicate that the model robustness is independent of the model size.

The results also show that our proposed attack method TSAttacker can generate high-quality adversarial samples because of the low average Levenshtein distance.
The average number of syllables in the datasets TNCC-title, TNCC-document, and TU\_SA is 16, 689, and 28.
For the dataset TNCC-title, the average Levenshtein distance of the generated adversarial samples on the models cino-base-v2 and cino-large-v2 is 1.6411 and 1.7176, respectively;
for the dataset TNCC-document, the average Levenshtein distance of the generated adversarial samples on the models cino-base-v2 and cino-large-v2 is 39.1800 and 41.9660, respectively;
for the dataset TU\_SA, the average Levenshtein distance of the generated adversarial samples on the models cino-base-v2 and cino-large-v2 is 2.9404 and 2.7017, respectively.
Several examples in Appendix \ref{AppendixB} intuitively demonstrate that the model's prediction transforms from one high-confidence classification to another after conducting TSAttacker.

\subsection{Ablation Experiment}

\begin{figure*}[h]
	\caption{Results of ablation experiments.}
	\label{Figure1}
	\centering
	\begin{subfigure}{0.49\linewidth}
		\centering
		\includegraphics[scale=0.4]{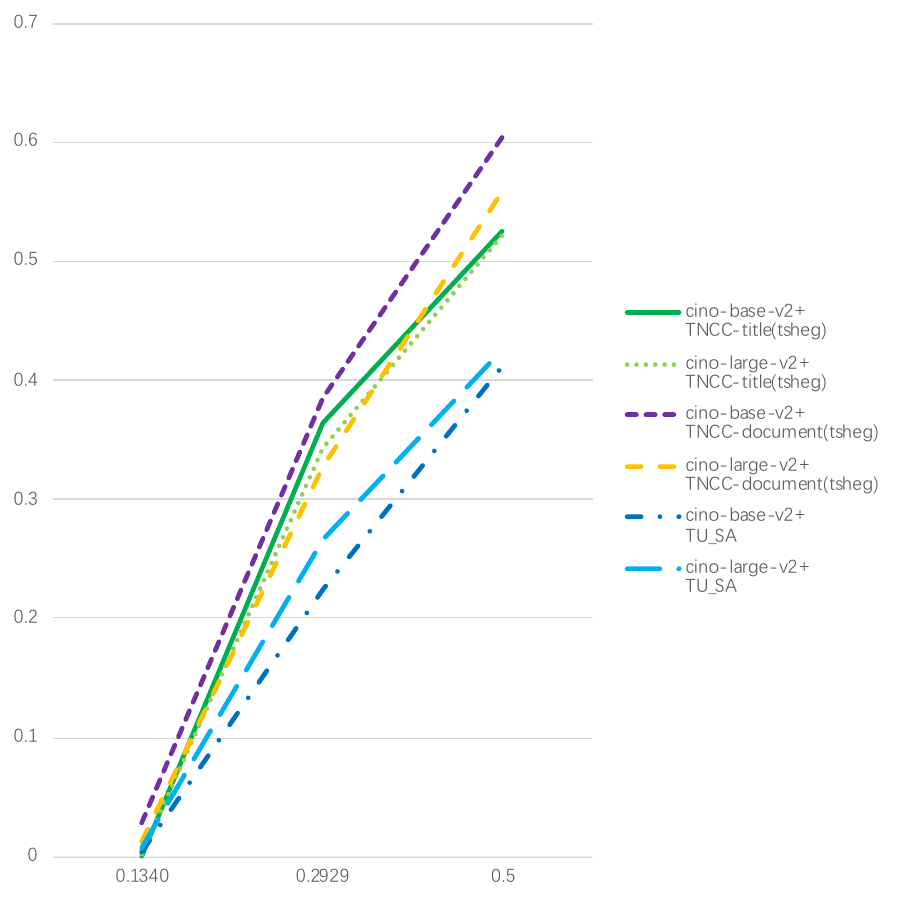}
		\caption{Accuracy Drop Value}
	\end{subfigure}
	\begin{subfigure}{0.49\linewidth}
		\centering
		\includegraphics[scale=0.4]{picture1.pdf}
		\caption{Attack Success Rate}
	\end{subfigure}
	\qquad
	\begin{subfigure}{\linewidth}
		\centering
		\begin{minipage}{0.49\linewidth}
			\centering
			\includegraphics[scale=0.4]{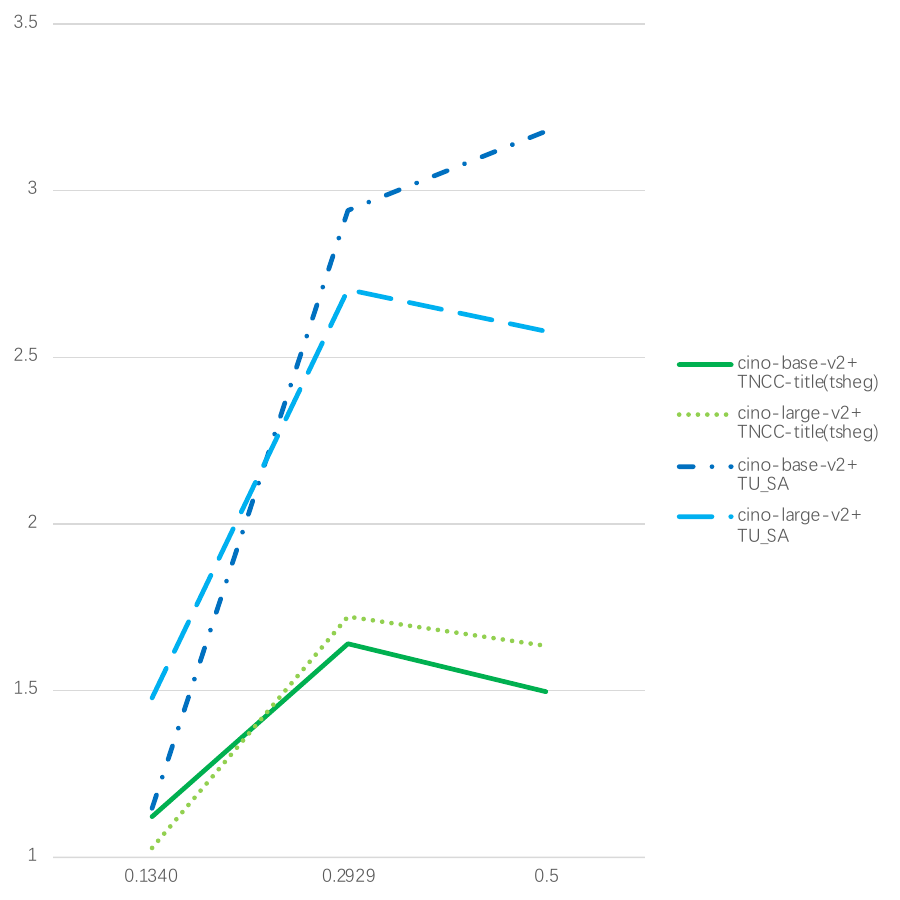}
		\end{minipage}
		\begin{minipage}{0.49\linewidth}
			\centering
			\includegraphics[scale=0.4]{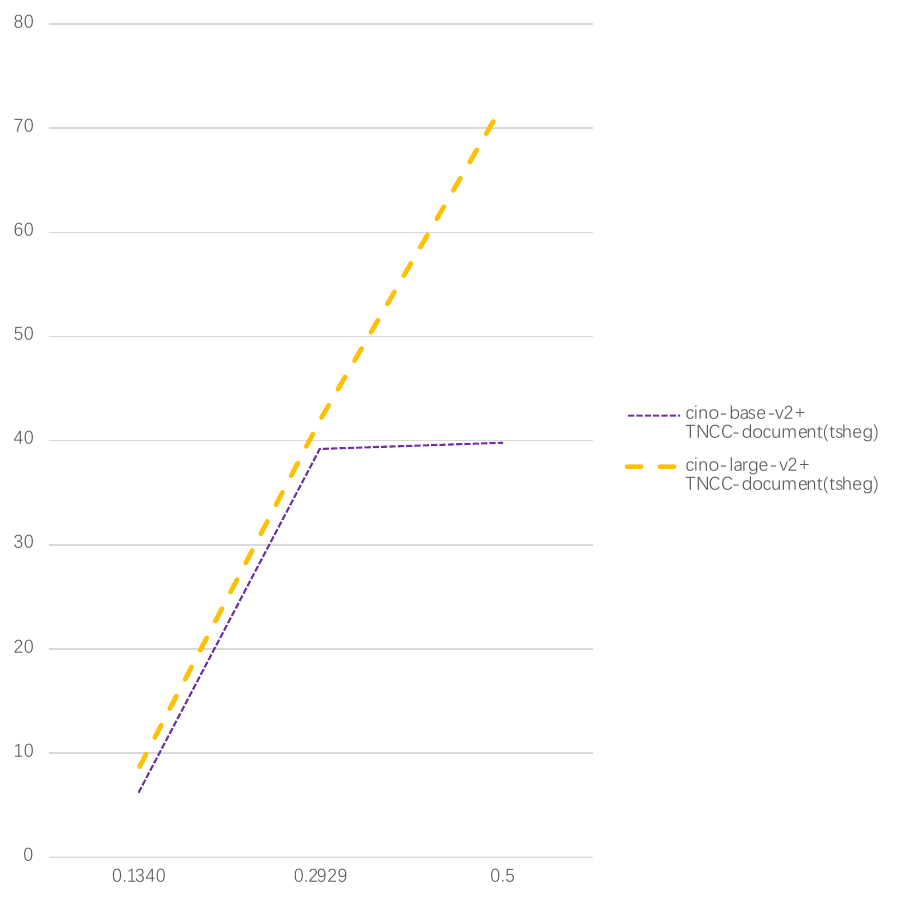}
		\end{minipage}
		\caption{Average Levenshtein Distance}
	\end{subfigure}
\end{figure*}

Since our experiments involve an artificially set parameter, the maximum cosine distance $d_{max}$, we explore the influence of $d_{max}$ on various evaluation metrics through ablation experiments as follows.
We set $d_{max}$ to 0.1340, 0.2929, and 0.5, respectively, that is to say, we set the maximum angle between two syllable embeddings to $30^{\circ}$, $45^{\circ}$, and $60^{\circ}$ to get the set of candidate substitution syllables, then we conduct TSAttacker on the six models.
Figure \ref{Figure1} shows the results of the ablation experiments in the form of line charts.
From the line charts, we can intuitively find that the larger $d_{max}$, the larger accuracy drop value and attack success rate, and the relationship between $d_{max}$ and average Levenshtein distance is not significant.
Although the larger $d_{max}$, the more effective the attack, the similarity between the substituted syllable and the substitution syllable may not be that high.

\section{Discussion}

\subsection{Textual Adversarial Attack is a Major Threat}

Recently, \citet{wang-etal-2023-on} evaluated the adversarial robustness of ChatGPT and found that the absolute performance of ChatGPT is far from perfection even though it outperforms most of the counterparts.
Nowadays, more and more applications based on the services of foundation models appear, making various downstream scenarios face the risk of textual adversarial attacks worryingly.
They also found that some small models achieve better performance on adversarial tasks while having much fewer parameters than the strong models.
Therefore, there is still great space for research on the robustness and interpretability of neural network models.

\subsection{Pay Attention to the Robustness of Chinese Minority Language Models}

The textual adversarial attack is a new challenge for Chinese minority languages' information processing, which poses a major threat to the stable development and information construction of Chinese minority areas.
China is a unified multi-ethnic country.
Due to the late start of information processing technology for Chinese minority languages, there is little research on the textual adversarial attack and defense of Chinese minority languages nowadays.
With the development of neural network models, research in this field is now urgent.

From an attack perspective.
The attack method proposed in this paper only preliminarily explores the field and evaluates the robustness of the Tibetan part in the first Chinese minority multilingual PLM.
Moreover, the attack methods combined with the linguistic characteristics of Chinese minority languages need to be further proposed.

From a defense perspective.
The overall performance of Chinese minority PLMs, including robustness, is far worse than that of English and Chinese PLMs.
The main reason is that there is a huge gap in the quantity level between the corpus of Chinese minority languages and the corpus of English and Chinese.
Therefore, this problem should be alleviated first.
In addition, in response to the proposed textual adversarial attacks, a posterior defense is also an effective method.

\section{Conclusion}

In this work, we propose a Tibetan syllable-level black-box textual adversarial attack called TSAttacker.
In TSAttacker, the syllable cosine distance is used to obtain syllables for substitution, and the scoring mechanism is used to determine the order of syllable substitutions.
We conduct TSAttacker on six models generated by fine-tuning two versions of the PLM CINO for three downstream tasks.
The experiment results show that TSAttacker greatly reduces the model accuracy and has a high attack success rate.
Also, the adversarial samples generated by TSAttacker are high-quality.
From a certain point of view, the robustness of the models still has much room for improvement.

\section*{Acknowledgments}
We would like to express our sincere gratitude to the following funding sources for their support of this article: the ``New Generation Artificial Intelligence'' Major Project of Science and Technology Innovation 2030 (No. 2022ZD0116100), the National Natural Science Foundation of China (No. 62162057), and the Mount Everest Discipline Construction Project of Tibet University (Project No. zf22002001).

\section*{Limitations}
%
Our work only preliminarily explores the field of textual adversarial attack on Chinese minority languages and evaluates the robustness of the Tibetan part in the first Chinese minority multilingual PLM.
The textual adversarial attack is a major threat in the information processing of Chinese minority languages.
We hope our attack method, experiment results, and discussions could provide experience for future research.
In the future, we will continue to concentrate on the security issues faced in Tibetan information processing.

\section*{Ethics Statement}
The purpose of this paper is to show that our proposed attack method is effective and the robustness of the SOTA Chinese minority PLM CINO still has much room for improvement, but not to attack it intentionally.
Finally, we call on more researchers to pay attention to the security issues in the information processing of Chinese minority languages.


\bibliography{trustnlp2023}
\bibliographystyle{acl_natbib}

\clearpage

\onecolumn

\appendix

%

\section{Pseudocode of TSAttacker Algorithm}
\label{AppendixA}

\begin{algorithm}[h]
	\caption{TSAttacker Algorithm}
	\label{Algorithm1}
	\KwIn{Classifier $F$.}
	\KwIn{Original text $x=s_{1}s_{2}\dots{s_{i}}\dots{s_{n}}$.}
	\KwIn{Maximum cosine distance $d_{max}$.}
	\KwOut{Adversarial text $x'$.}
	\For{$i\leftarrow1$ to $n$}
	{
		$\hat{{x}_{i}}\leftarrow{s_{1}s_{2}\dots{<UNK>}\dots{s_{n}}}$\tcp*[f]{Equation \ref{Equation10}} \\
		$S_{i}\leftarrow{P(y_{true}|x)-P(y_{true}|\hat{x_{i}})}$\tcp*[f]{Equation \ref{Equation11}}
	}
	Init $H$ as a empty list. \\
	\For{$i\leftarrow1$ to $n$}
	{
		Get the candidate syllables’ set $C_{i}$ according to syllable $s_{i}$ and $d_{max}$. \\
		$m\leftarrow{len(C_{i})}$ \\
		\For{$j\leftarrow1$ to $m$}
		{
			${s_{i}}'\leftarrow{C_{ij}}$ \\
			{${x_{i}}'\leftarrow{s_{1}s_{2}\dots{{s_{i}}'}\dots{s_{n}}}$}\tcp*[f]{Equation \ref{Equation5}} \\
			$\Delta{P_{i}}\leftarrow{P(y_{true}|x)-P(y_{true}|{x_{i}}')}$\tcp*[f]{Equation \ref{Equation6}}
		}
		$\Delta{{P_{i}}^{*}}\leftarrow{max\{\Delta{P_{ij}\}}_{j=1}^{m}}$\tcp*[f]{Equation \ref{Equation8}} \\
		${s_{i}}^{*}\leftarrow{arg\max_{{{s_{i}}'}\in{C_{i}}}\{\Delta{P_{ij}\}}_{j=1}^{m}}$\tcp*[f]{Equation \ref{Equation9}} \\
		$H_{i}\leftarrow{\frac{e^{S_{i}}}{\Sigma_{j=1}^{n}e^{S_{j}}}\cdot\Delta{{P_{i}}^{*}}}$\tcp*[f]{Equation \ref{Equation12}} \\
		Append $({s_{i}}^{*},H_{i})$ into $H$.
	}
	Sort $H$ by the second parameter in descending order. \\
	\ForEach{element in $H$}
	{
		$x'\leftarrow{s_{1}s_{2}\dots{{s_{i}}^{*}}\dots{s_{n}}}$ \\
		\If{${F(x')}\neq{F(x)}$}
		{
			Attack succeeds and return $x'$.
		}
	}
	Attack fails and return.
\end{algorithm}

\clearpage

\section{Some Generated Adversarial Samples}
\label{AppendixB}

\begin{table}[h]
	\centering
	\begin{tabular}{m{2.5cm}<{\centering}m{8cm}<{\centering}m{2cm}<{\centering}m{2cm}<{\centering}}
		\toprule
		Model & Input & \makecell{Output\\(pre-attack)} & \makecell{Output\\(post-attack)} \\
		\midrule
		\makecell{cino-large-v2+\\TNCC-title\\(tsheg)} & \includegraphics[scale=0.35]{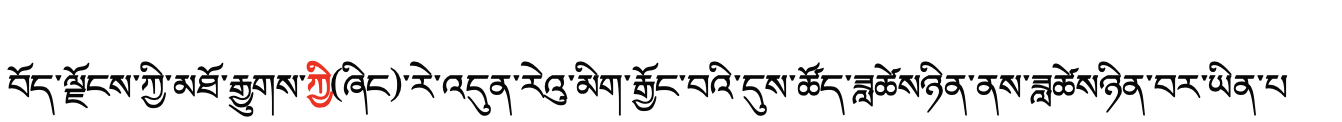} & \makecell{Education\\(92.95\%)} & \makecell{Economics\\(97.35\%)} \\
		\makecell{cino-large-v2+\\TNCC-document\\(tsheg)} & \includegraphics[scale=0.35]{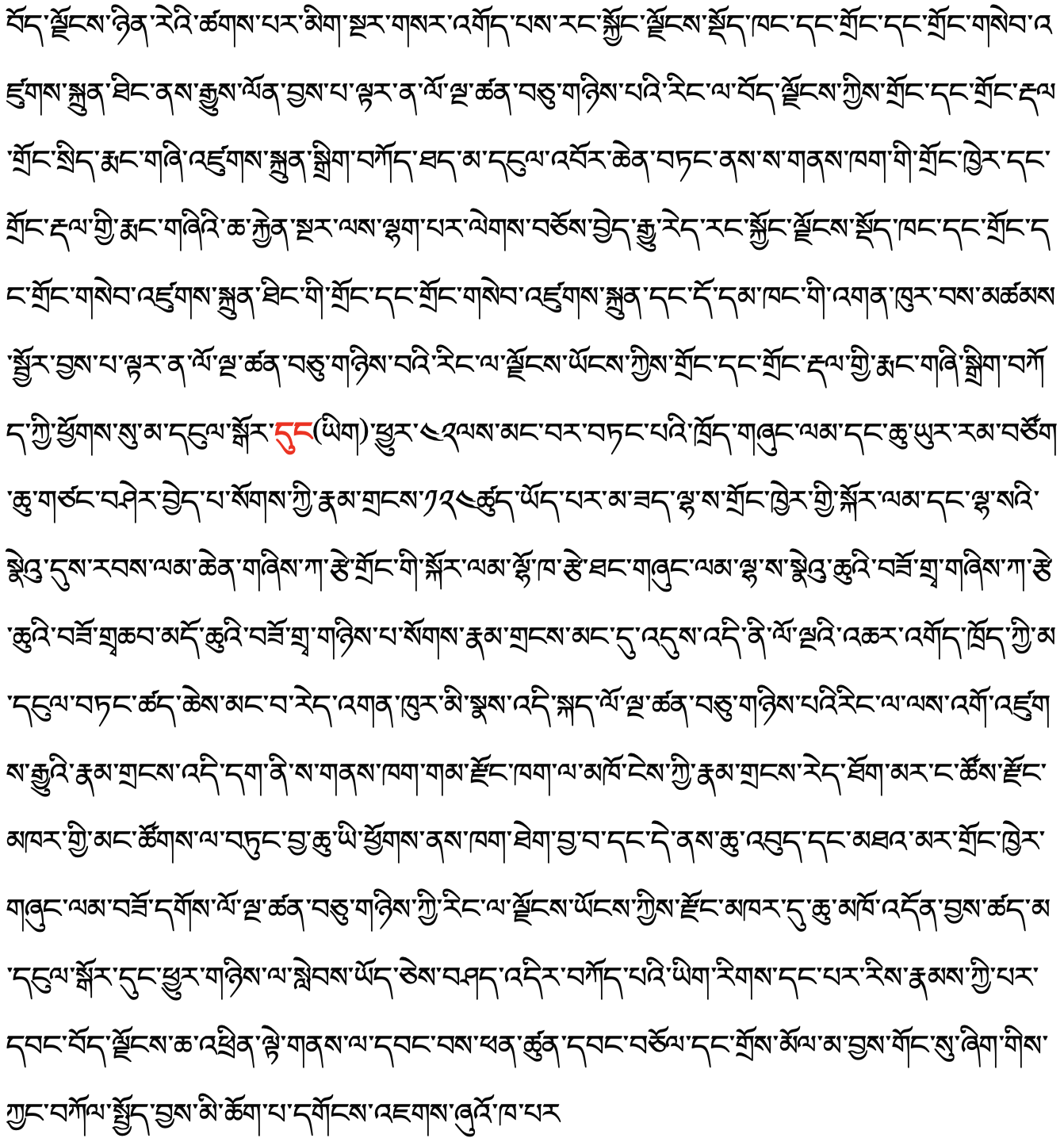} & \makecell{Economics\\(91.24\%)} & \makecell{Environment\\(99.86\%)} \\
		\makecell{cino-large-v2+\\TU\_SA} & \includegraphics[scale=0.35]{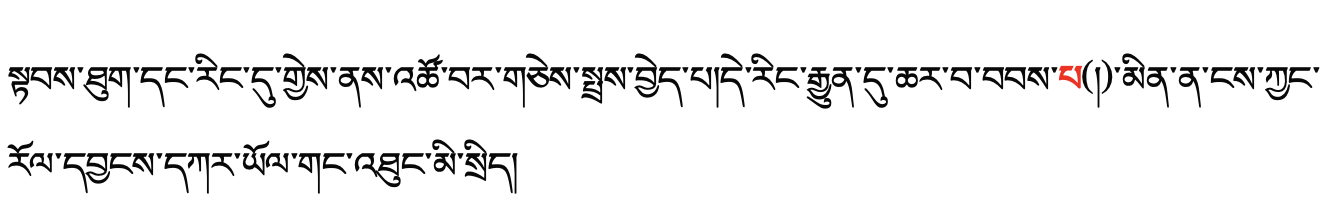} & \makecell{Negative\\(94.74\%)} & \makecell{Positive\\(99.91\%)} \\
		\bottomrule
		 &  &  &  \\
	\end{tabular}
	Substituted syllables are marked in bold and red. Substitution syllables are in parentheses.
\end{table}

\end{document}